\documentclass[twoside]{article}
\usepackage[accepted]{aistats2018}
\usepackage[utf8]{inputenc} 
\usepackage[T1]{fontenc}    
\usepackage{hyperref}       
\usepackage{url}            
\usepackage{booktabs}       
\usepackage{amsfonts}       
\usepackage{nicefrac}       
\usepackage{microtype}      

\usepackage{tikz}

\usepackage{fancyvrb}
\usepackage{authblk}
\usepackage{hyperref}
\usepackage{amssymb}
\usepackage{amsmath}
\usepackage{amsthm}
\usepackage{algorithm,algpseudocode}
\algtext*{EndWhile}
\algtext*{EndIf}
\usepackage{enumerate}
\usepackage{xcolor}
\usepackage{graphicx}
\newcommand{\myfootnote}[1]{
	\renewcommand{\thefootnote}{}
	\footnotetext{\hspace{-16.5pt}\small#1}
	\renewcommand{\thefootnote}{\arabic{footnote}}
}
\newcommand{\mytilde}{\raise.17ex\hbox{$\scriptstyle\mathtt{\sim}$}}

\DeclareMathOperator*{\argmin}{arg\,min}
\DeclareMathOperator{\E}{\mathbb{E}}
\newtheorem{de}{Definition}
\newtheorem{theo}{Theorem}

\newtheorem*{rem}{Remark}
\newcommand{\R}{\mathbb{R}}

\newtheorem{innercustomthm}{Theorem}
\newenvironment{customthm}[1]
  {\renewcommand\theinnercustomthm{#1}\innercustomthm}
  {\endinnercustomthm}

\begin{document}

%

%

\twocolumn[

\aistatstitle{Multi-view Metric Learning in Vector-valued Kernel Spaces}

\aistatsauthor{ Riikka Huusari \And Hachem Kadri \And  Cécile Capponi }
\vspace*{0.3cm}
\aistatsaddress{ Aix Marseille Univ, Université de Toulon, CNRS, LIS, Marseille, France  } ]

\begin{abstract}
We consider the problem of metric learning for multi-view data and present a novel method for learning within-view as well as between-view metrics in vector-valued kernel spaces, as a way to capture multi-modal structure of the data. 
We formulate two convex optimization problems to jointly learn the metric and the classifier or regressor in kernel feature spaces. 
An iterative three-step multi-view metric learning algorithm is derived from the optimization problems. 
In order to scale the computation to large training sets, a block-wise Nyström approximation of the multi-view kernel matrix is introduced. 
We justify our approach theoretically and experimentally, and show its performance on real-world datasets against relevant state-of-the-art methods.
\end{abstract}

\section{Introduction}


In this paper we tackle the problem of supervised multi-view learning, where each labeled example is observed under several views.
These views might be not only correlated, but also complementary, redundant or contradictory.  
Thus, learning over all the views is expected to produce a final classifier (or regressor) that is better than each individual one. 
Multi-view learning is well-known in the semi-supervised setting, where the agreement among views is usually optimized \cite{Blum98combininglabeled,Sindhwani08rkhs}.
Yet, the supervised setting has proven to be interesting as well, independently from any agreement condition on views. 
Co-regularization and multiple kernel learning~(MKL) are two well known kernel-based frameworks for learning in the presence of multiple views of data~\cite{xu2013survey}. The former attempts to optimize measures of agreement and smoothness between the views over labeled and unlabeled examples~\cite{sindhwani2005co}; the latter tries to efficiently combine multiple kernels defined on each view to exploit information coming from different representations~\cite{gonen2011multiple}.
More recently, vector-valued reproducing kernel Hilbert spaces~(RKHSs) have been introduced to the field of multi-view learning for going further than MKL by incorporating in the learning model both within-view and between-view dependencies~\cite{Minh2013unifying, Kadri2013multi}.
It turns out that these kernels and their associated vector-valued reproducing  Hilbert spaces provide a unifying framework for a number of previous multi-view kernel methods, such as co-regularized multi-view learning and manifold regularization, and naturally allow to encode within-view as well as between-view similarities~\cite{Minh2016unifying}.

Kernels of vector-valued RKHSs are positive semidefinite matrix-valued functions. They have been applied with success in various machine learning problems, such as multi-task learning~\cite{Evgeniou2005learning}, functional regression~\cite{Kadri2015operator} and structured output prediction~\cite{Brouard2016input}. 
The main advantage of matrix-valued kernels is that they offer a higher degree of flexibility in encoding similarities between data points.
However finding the optimal matrix-valued kernel of choice for a given application is difficult, as is the question of how to build them.
In order to overcome the need for choosing a kernel before the learning process, we propose a supervised metric learning approach that learns a matrix-valued multi-view kernel jointly with the decision function.
We refer the reader to~\cite{bellet2015metric} for a review of metric learning.
It is worth mentioning that algorithms for learning matrix-valued kernels have been proposed in the literature, see for example~\cite{dinuzzo2011learning, Ciliberto2015convex, lim2015operator}. However, these methods mainly consider separable kernels which are not suited for multi-view setting, as will be illustrated later in this paper.

The main contributions of this paper are: 1) we introduce and learn a new class of matrix-valued kernels designed to handle multi-view data 2) we give an iterative algorithm that learns simultaneously a vector-valued multi-view function and a block-structured metric between views, 3) we provide generalization analysis of our algorithm with a Rademacher bound; and 4) we show how matrix-valued kernels can be efficiently computed via a block-wise Nyström approximation in order to reduce significantly their high computational cost.

\section{Preliminaries}

We start here by briefly reviewing the basics of vector-valued RKHSs and their associated matrix-valued kernels. We then describe how they can be used for learning from multi-view data.

\subsection{Vector-valued RKHSs}

Vector-valued RKHSs were introduced to the machine learning community by Micchelli and Pontil~\cite{Micchelli2005onlearning} as a way to extend kernel machines from scalar to vector outputs.
In this setting, given a random training sample $\{\mathbf{x}_i,y_i\}_{i=1}^n$ on $\mathcal{X} \times \mathcal{Y}$, optimization problem 
 \begin{equation}
 \label{eq:minRKHS}
\arg\min_{f\in\mathcal{H}} \sum_{i=1}^n V(f,\mathbf{x}_i,y_i) + \lambda \|f\|_\mathcal{H}^2,
 \end{equation}
where $f$ is a vector-valued function and $V$ is a loss function, can be solved in a vector-valued RKHS $\mathcal{H}$ by the means of a vector-valued extension of the representer theorem.
To see this more clearly, we recall some fundamentals of vector-valued RKHSs. 
\vspace{-0.0cm}
\begin{de} (vector-valued RKHS) \\[0.1cm] 
  A Hilbert space $\mathcal{H}$ of functions from $\mathcal{X}$ to
  $\mathbb{R}^v$ is called a reproducing kernel Hilbert space if
  there is a positive definite $\mathbb{R}^{v\times v}$-valued kernel
  $K$ on $\mathcal{X} \times \mathcal{X}$ such that:
\vspace{-0.3cm}
  \begin{enumerate}[i.]
    \item \label{enum1:i} the function $z \mapsto K(\mathbf{x},\mathbf{z})\mathbf{y}$ belongs to $\mathcal{H},\ \forall \mathbf{z}, \mathbf{x} \in \mathcal{X},\ \mathbf{y} \in \mathbb{R}^{v}$,
    \item \label{enum1:ii} $\forall f \in \mathcal{H}, \mathbf{x} \in \mathcal{X},\ \mathbf{y} \in \mathbb{R}^{v}, \ \ 
      \langle f,K(\mathbf{x},\cdot)\mathbf{y}\rangle _{\mathcal{H}} =
      \langle f(\mathbf{x}),\mathbf{y}\rangle_{\mathbb{R}^{v}}$ \hspace*{0.1cm} (reproducing property).
  \end{enumerate}
\end{de}
\begin{de} (matrix-valued kernel) \\[0.1cm]
  An $\mathbb{R}^{v\times v}$-valued kernel $K$ on
  $\mathcal{X}\!\!\;\times\!\!\;\mathcal{X}$ is a function
  $K(\cdot,\cdot):\mathcal{X} \times \mathcal{X}
  \rightarrow \mathbb{R}^{v\times v}$; it is positive semidefinite if:
\vspace{-0.3cm}
  \begin{enumerate}[i.]
    \item $K(\mathbf{x}, \mathbf{z})=K(\mathbf{z}, \mathbf{x})^{\top}$, where $^\top$ denotes the transpose of a matrix, 
    \item  and, for every $r\in\mathbb{N}$ and all
      $\{(\mathbf{x}_{i},y_{i})_{i=1,\ldots ,r}\}\in \mathcal{X} \times
      \mathbb{R}^{v}$,  $\sum_{i,j} \langle
      y_i, K(\mathbf{x}_{i},\mathbf{x}_{j})y_{j}\rangle_{\mathbb{R}^{v}} \geq 0$.
  \end{enumerate}
\end{de}

Important results for matrix-valued kernels include the positive semidefiniteness of the kernel $K$ and that we obtain a solution for regularized optimization problem~(\ref{eq:minRKHS}) via a representer theorem. It states that solution $\hat{f}\in \mathcal{H}$ for a learning problem can be written as 
\[\hat{f}(\mathbf{x}) = \sum_{i=1}^n K(\mathbf{x},\mathbf{x}_i)\mathbf{c}_i,\;\;\; \text{with} \;\;\; \mathbf{c}_i\in \mathbb{R}^v.\] 

Some well-known classes of matrix-valued kernels include separable and transformable kernels. Separable kernels are defined by 
\[K(\mathbf{x}, \mathbf{z}) = k(\mathbf{x}, \mathbf{z})\mathbf{T},\] where $\mathbf{T}$ is a matrix in $\mathbb{R}^{v\times v}$. This class of kernels is very attractive in terms of computational time, as it is easily decomposable. However the matrix $\mathbf{T}$ acts only on the outputs independently of the input data, which makes it difficult for these kernels to encode necessary similarities in multi-view setting. Transformable kernels are defined by  
\[[K(\mathbf{x}, \mathbf{z})]_{lm} = k(S_m\mathbf{x}, S_l\mathbf{z}).\] 
Here $m$ and $l$ are indices of the output matrix~(views in multi-view setting) and operators $\{S_t\}_{t=1}^v$, are used to transform the data. In contrast to separable kernels, here the $S_t$ operate on input data; however choosing them is a difficult task.  
For further reading on matrix-valued reproducing kernels, see, e.g.,~\cite{alvarez2012kernels,Caponnetto2008universal, Carmeli2010vector, Kadri2015operator}.

\subsection{Vector-valued multi-view learning}

This section reviews the setup for supervised multi-view learning in vector-valued RKHSs~\cite{Kadri2013multi, Minh2016unifying}.  
The main idea is to consider a kernel that measures not only the similarities between examples of the same view but also those coming from different views. 
Reproducing kernels of vector-valued Hilbert spaces allow encoding in a natural way these similarities and taking into account both within-view and between-view dependencies. 
Indeed, a kernel function $K$ in this setting outputs a matrix in $\mathbb{R}^{v\times v}$, with $v$ the number of views, 
so, that $K(\mathbf{x}_i,\mathbf{x}_j)_{lm}$, $l,m = 1,\ldots,v$, is the similarity measure between examples $\mathbf{x}_i$ and $\mathbf{x}_j$ from the views $l$ and $m$. 

More formally, consider a set of $n$ labeled data $\{(\mathbf{x}_i,y_i) \in \mathcal{X}\times \mathcal{Y}, i =1,\ldots,n\}$, where $\mathcal{X} \subset\mathbb{R}^d$ and $\mathcal{Y}=\{-1,1\}$ for  classification or $\mathcal{Y} \subset \mathbb{R}$ for regression. Also assume that each input instance $\mathbf{x}_i = (\mathbf{x}_i^1, \ldots, \mathbf{x}_i^v)$ is seen in $v$ views, where $x_i^l \in \mathbb{R}^{d_l}$ and $\sum_{l=1}^v d_l = d$. The supervised multi-view learning problem can be thought of as trying to find the vector-valued function $\hat{f}(\cdot) = (\hat{f}^1(\cdot),\ldots \hat{f}^v(\cdot))$, with $\hat{f}^l(\mathbf{x}) \in \mathcal{Y}$, solution of 
\begin{equation}
\label{eq:vvmvl}
\argmin_{f\in\mathcal{H}, \mathcal{W} } \sum_{i=1}^n V(y_i, \mathcal{W}(f(\mathbf{x}_i))) + \lambda \|f\|^2.
\end{equation}
Here $f$ is a vector-valued function that groups $v$ learning functions, each corresponding to one view, and $\mathcal{W}:\R^v \rightarrow \R$ is combination operator for combining the results of the learning functions.

While the vector-valued extension of the representer theorem provides an algorithmc way for computing the solution of the multi-view learning problem~(\ref{eq:vvmvl}), the question of choosing the multi-view kernel $K$ remains crucial to take full advantage of the vector-valued learning framework. 
In~\cite{Kadri2013multi}, a matrix-valued kernel based on cross-covariance operators on RKHS that allow modeling variables of multiple types and modalities was proposed. However, it has two major drawbacks: i) the kernel is fixed in advance and does not depend on the learning problem, and ii) it is computationally expensive and becomes infeasible when
the problem size is very large. We avoid both of these issues by learning a block low-rank metric in kernel feature spaces.

\section{Multi-View Metric Learning}

Here we introduce an optimization problem for learning simultaneously a vector-valued multi-view function and a positive semidefinite metric between kernel feature maps, as well as an operator for combining the answers from the views to yield the final decision. We then derive a three-step metric learning algorithm for multi-view data and give Rademacher bound for it. Finally we demonstrate how it can be implemented efficiently via block-wise Nyström approximation and give a block-sparse version of our formulation.

\subsection{Matrix-valued multi-view kernel}
We consider the following class of matrix-valued kernels that can operate over multiple views
\begin{equation}
\label{eq:mvk}
K(\mathbf{x}_i,\mathbf{x}_j)_{lm} = \left\langle \Phi_l(\mathbf{x}_i^l), C_{\mathcal{X}_l\mathcal{X}_m} \Phi_m(\mathbf{x}_j^m) \right\rangle, 
\end{equation}
where $\Phi_l$ (resp. $\Phi_m$) is the feature map associated to the scalar-valued kernel $k_l$ (resp. $k_m$) defined on the view $l$ (resp. $m$). In the following we will leave out the view label from data instance when the feature map or kernel function already has that information, e.g. instead of $ \Phi_l(\mathbf{x}_i^l)$ we write $ \Phi_l(\mathbf{x}_i)$.
$C_{\mathcal{X}_l\mathcal{X}_m} : \mathcal{H}_m \to \mathcal{H}_l$ is a linear operator between the scalar-valued RKHSs $\mathcal{H}_l$ and $\mathcal{H}_m$ of kernels $k_l$ and $k_m$, respectively. 
The operator $C_{\mathcal{X}_l\mathcal{X}_m}$ allows one to encode both within-view and between-view similarities. 

The choice of the operator $C_{\mathcal{X}_l\mathcal{X}_m}$  is crucial and depends on the multi-view problem at hand. 
In the following we only consider operators $C_{\mathcal{X}_l\mathcal{X}_m} $ that can be written as $C_{\mathcal{X}_l\mathcal{X}_m} = \mathbf{\Phi}_l \mathbf{A}_{lm} \mathbf{\Phi}^T_m$, where $\mathbf{\Phi}_s = \left(\Phi_s(\mathbf{x}_1),...,\Phi_s(\mathbf{x}_n)\right)$ with $s=l,m$ 
and $\mathbf{A}_{lm} \in \mathbb{R}^{n\times n}$ is a positive definite matrix which plays the role of a metric between the two features maps associated with kernels $k_l$ and $k_m$ defined over the views $l$ and $m$. 
This is a large set of possible operators, but depends on a finite number of parameters. 
It gives us the following class of kernels
\begin{align}\label{eq:mvka}
K(\mathbf{x}_i,\mathbf{x}_j)_{lm} &= \left\langle \Phi_l(\mathbf{x}_i), \mathbf{\Phi}_l \mathbf{A}_{lm} \mathbf{\Phi}^T_m \Phi_m(\mathbf{x}_j) \right\rangle \notag \\
&= \left\langle \mathbf{\Phi}_l^T \Phi_l(\mathbf{x}_i),  \mathbf{A}_{lm} \mathbf{\Phi}^T_m \Phi_m(\mathbf{x}_j) \right\rangle  \notag \\
&= \left\langle \mathbf{k}_l(\mathbf{x}_i),  \mathbf{A}_{lm} \mathbf{k}_m(\mathbf{x}_i) \right\rangle, 
\end{align}
where we have written $\mathbf{k}_l(\mathbf{x}_i) = (k_l(\mathbf{x}_t, \mathbf{x}_i))_{t=1}^n$. We note that this class is not in general separable or transformable. However in the special case when it is possible to write $\mathbf{A}_{ml} = \mathbf{A}_m\mathbf{A}_l$ the kernel is transformable.

It is easy to see that the $lm$-th block of the block kernel matrix $\mathbf{K}$ built from the matrix-valued kernel~(\ref{eq:mvka}) can be written as
$\mathbf{K}_{lm} = \mathbf{K}_l\mathbf{A}_{lm}\mathbf{K}_m,$
where $\mathbf{K}_s = \big(k_s(\mathbf{x}_i,\mathbf{x}_j)\big)_{i,j=1}^n$ for view $s$. 
The block kernel matrix $\mathbf{K} = \big(K(\mathbf{x}_i,\mathbf{x}_j)\big)_{i,j=1}^n$ in this case has the form
\begin{equation}
\label{eq:mvkm}
\mathbf{K} = \mathbf{H}\mathbf{A}\mathbf{H},
\end{equation}
where $\mathbf{H} = \text{blockdiag}(\mathbf{K}_1,\cdots, \mathbf{K}_v)$,\footnote{Given a set of $n\times n $ matrices $\mathbf{K}_1,\cdots, \mathbf{K}_v$, $\mathbf{H} = \text{blockdiag}(\mathbf{K}_1,\cdots, \mathbf{K}_v)$ is the block diagonal matrix satisfying $\mathbf{H}_{l,l} = \mathbf{K}_l, \forall l=1,\ldots,v$.} and the matrix $\mathbf{A} = (\mathbf{A}_{lm})_{l,m=1}^v \in \mathbb{R}^{nv\times nv}$ encodes pairwise similarities between all the views. Multi-view metric learning then corresponds to simultaneously learning the metric~$\mathbf{A}$ and the classifier or regressor.

From this framework, with suitable choices of $\mathbf{A}$, we can recover the cross-covariance multi-view kernel of~\cite{Kadri2013multi}, or for example a MKL-like multi-view kernel containing only one-view kernels.

\subsection{Algorithm}
\label{subsec:algo}

Using the vector-valued representer theorem, the multi-view learning problem~(\ref{eq:vvmvl}) becomes
\begin{align*}
\argmin_{\mathbf{c}_1,\ldots, \mathbf{c}_n \in \mathbb{R}^{v}} &\sum_{i=1}^n V\left(y_i, \mathcal{W}\left(\sum_{j=1}^n K(\mathbf{x}_i,\mathbf{x}_j)\mathbf{c}_j\right)\right) \\ 
&+ \lambda \sum_{i,j=1}^n\langle \mathbf{c}_i, K(\mathbf{x}_i,\mathbf{x}_j) \mathbf{c}_j \rangle.
\end{align*}
We set $V$  to be the square loss function and assume the operator $\mathcal{W}$ to be known. We choose it to be weighted sum of the outputs: $\mathcal{W}=\mathbf{w}^T \otimes \mathbf{I}_n$ giving us $\mathcal{W}f(\mathbf{x}) = \sum_{m=1}^v w_mf^m(\mathbf{x})$. Let $\mathbf{y} \in \mathbb{R}^{n}$ be the output vector $(y_{i})_{i=1}^n$. The previous optimization problem can now be written as
\begin{equation*}
\arg\min_{\mathbf{c}\in\mathbb{R}^{nv}} \;\; \| \mathbf{y}- (\mathbf{w}^T \otimes \mathbf{I}_n)\mathbf{Kc}\|^2 + \lambda\left\langle \mathbf{Kc},\mathbf{c} \right\rangle,
\end{equation*}
where $\mathbf{K}\in \mathbb{R}^{nv\times nv}$ is the block kernel matrix associated to the matrix-valued kernel~(\ref{eq:mvk}) and $\mathbf{w}\in\R^v$ is a vector containing weights for combining the final result.

Using ~(\ref{eq:mvkm}), and considering an additional regularizer we formulate the multi-view metric learning~(MVML) optimization problem: 
\begin{align}
\label{eq:mvmlA}
 \min_{\mathbf{A},\mathbf{c}} \;\; &\| \mathbf{y}- (\mathbf{w}^T \otimes \mathbf{I}_n)\mathbf{H}\mathbf{AHc}\|^2 + \lambda\left\langle \mathbf{H}\mathbf{AHc},\mathbf{c} \right\rangle \\
&+ \eta\|\mathbf{A}\|_F^2, \;\;\;\; s.t. \;\; \mathbf{A} \succ 0. \notag
\end{align} 
Here we have restricted the block metric matrix $\mathbf{A}$ to be positive definite and we penalize its complexity via Frobenius norm.

Inspired by~\cite{Ciliberto2015convex} we make a change of variable $\mathbf{g} = \mathbf{AHc}$ in order to obtain a solution. Using a mapping $(\mathbf{c},\mathbf{A})\rightarrow (\mathbf{g},\mathbf{A})$ we obtain the equivalent learning problem: \begin{align}
\label{eq:mvmlbis}
\min_{\mathbf{A},\mathbf{g}} \;\; &\| \mathbf{y}- (\mathbf{w}^T \otimes \mathbf{I}_n)\mathbf{Hg}\|^2 + \lambda\left\langle \mathbf{g},\mathbf{A}^\dagger \mathbf{g} \right\rangle \\
&+ \eta\|\mathbf{A}\|_F^2, \;\;\;\; s.t. \;\; \mathbf{A} \succ 0. \notag
\end{align}
It is good to note that despite the misleading similarities between our work and that of~\cite{Ciliberto2015convex}, we use different mappings for solving our problems, which are also formulated differently. We also consider different classes of kernels as~\cite{Ciliberto2015convex} considers only separable kernels. 
\begin{rem}
The optimization problem~(\ref{eq:mvmlbis}) is convex. The main idea is to note that $\left\langle \mathbf{g},\mathbf{A}^\dagger \mathbf{g} \right\rangle$ is jointly convex~(see e.g.~\cite{Zhang2012convex}).
\end{rem}
We use an alternating scheme to solve our problem.  
We arrive to the following solution for $\mathbf{g}$ with fixed $\mathbf{A}$: 
\begin{equation}
\label{eq:D}
\mathbf{g} = (\mathbf{H}(\mathbf{w}^T \otimes \mathbf{I}_n)^T(\mathbf{w}^T \otimes \mathbf{I}_n)\mathbf{H}+\lambda \mathbf{A}^\dagger)^{-1}\mathbf{H}(\mathbf{w}^T \otimes \mathbf{I}_n)^T\mathbf{y}.
\end{equation}
The solution of~(\ref{eq:mvmlbis}) for $\mathbf{A}$ for fixed $\mathbf{g}$ is obtained by gradient descent, where the update rule is given by
\begin{equation}
\label{eq:A} 
\mathbf{A}^{k+1} = \left(1-2\mu\eta\right)\mathbf{A}^k + \mu\lambda\left(\mathbf{A}^k\right)^\dagger \mathbf{gg}^T\left(\mathbf{A}^k\right)^\dagger,
\end{equation}
where $\mu$ is the step size.
Technical details of the derivations can be found in the supplementary material (Appendix~A.1). It is important to note here that Equation~(\ref{eq:A}) is obtained by solving the optimization problem~(\ref{eq:mvmlbis}) without considering the positivity constraint on $\mathbf{A}$. Despite this, (when $\mu\eta<\tfrac{1}{2}$) the obtained $\mathbf{A}$ is symmetric and positive (compare to \cite{jawanpuria2015efficient}), and hence the learned matrix-valued multi-view kernel is valid.

If so desired, it is also possible to learn the weights $\textbf{w}$. For
fixed $\mathbf{g}$ and $\mathbf{A}$ the solution for $\mathbf{w}$ is   
\begin{equation}
\label{eq:w}
\mathbf{w} = (\mathbf{Z}^T\mathbf{Z})^{-1}\mathbf{Z}^T\mathbf{y},
\end{equation} 
where $\mathbf{Z}\in \R^{n\times v}$ is filled columnwise from $\mathbf{Hg}$. 

Our MVML algorithm thus iterates over solving $\mathbf{A}$, $\mathbf{g}$ and $\mathbf{w}$ if weights are to be learned (see Algorithm~\ref{alg:mvml}, version a).
The complexity of the algorithm is $\mathcal{O}(v^3n^3)$ for it computes the inverse of the $nv\times nv$ matrix $\mathbf{A}$, required for calculating $\mathbf{g}$. We will show later how to reduce the computational complexity of our algorithm via Nystr\"{o}m approximation, while conserving the desirable information about the multi-view problem.

\begin{algorithm}[t]
\caption{Multi-View Metric Learning: with a) full kernel matrices; b) Nyström approximation}\label{alg:mvml}
\begin{algorithmic}
\State {\textbf{Initialize} $\mathbf{A} \succ 0 $ and $\mathbf{w}$}

\While {not converged}
    \State {Update $\mathbf{g}$ via Equation a) \ref{eq:D} or b) \ref{eq:D_approx}}
    \If {$\mathbf{w}$ is to be calculated}
      \State {Update $\mathbf{w}$ via Equation a) \ref{eq:w} or b) \ref{eq:w_approx}}
    \EndIf
    \If {sparsity promoting method}
      \State {Iterate $\mathbf{A}$ with Equation a) \ref{eq:A_iter} or b) \ref{eq:A_approx_iter}}
    \Else
      \State {Iterate $\mathbf{A}$ via Equation a) \ref{eq:A} or b) \ref{eq:A_approx}}
    \EndIf
\EndWhile \\
\Return {$\mathbf{A}$, $\mathbf{g}$, $\mathbf{w}$}
\end{algorithmic}
\end{algorithm}

\begin{figure*}[!htbp]
\centering
\includegraphics[width=0.23\linewidth]{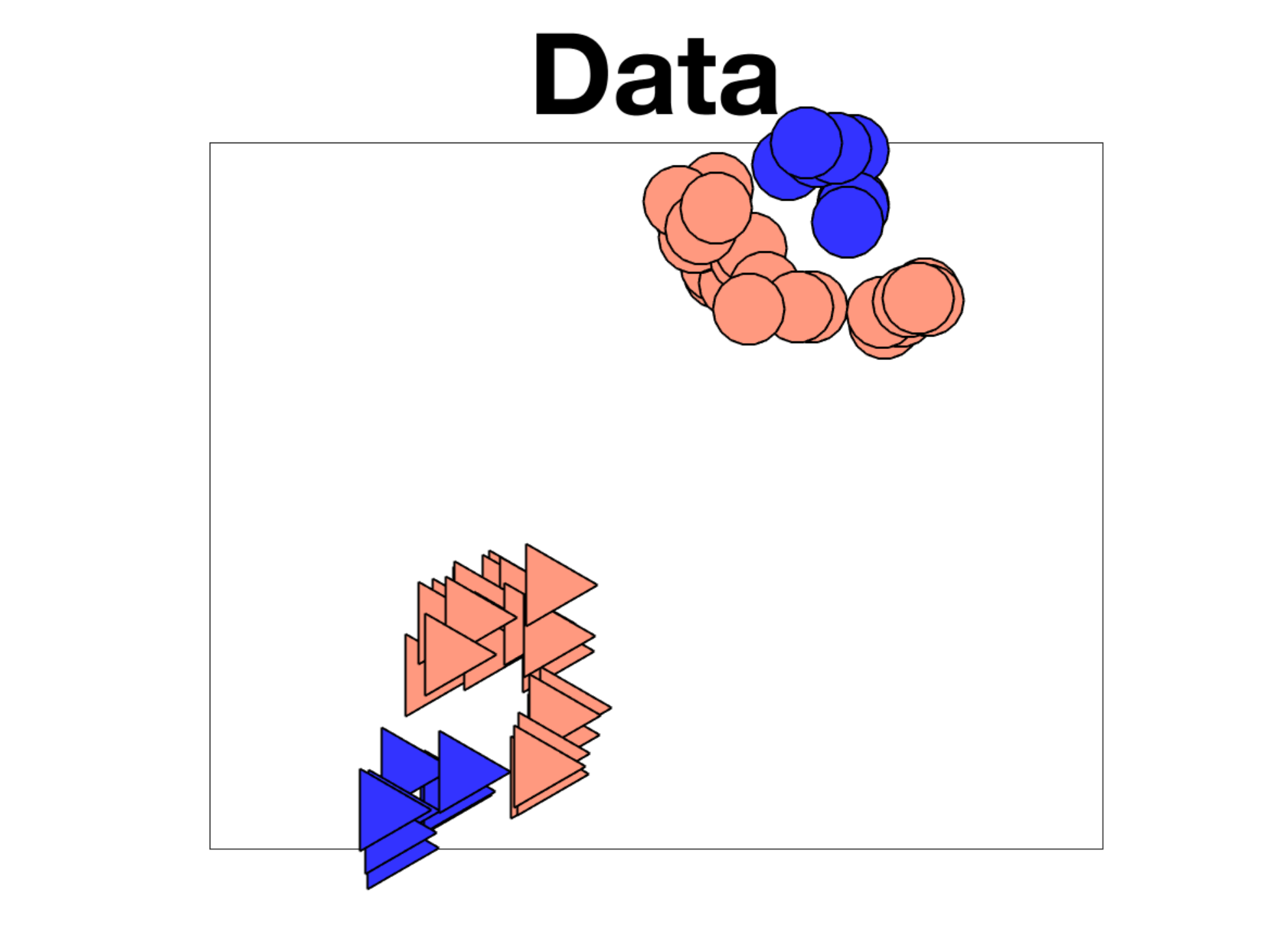}
\includegraphics[width=0.23\linewidth]{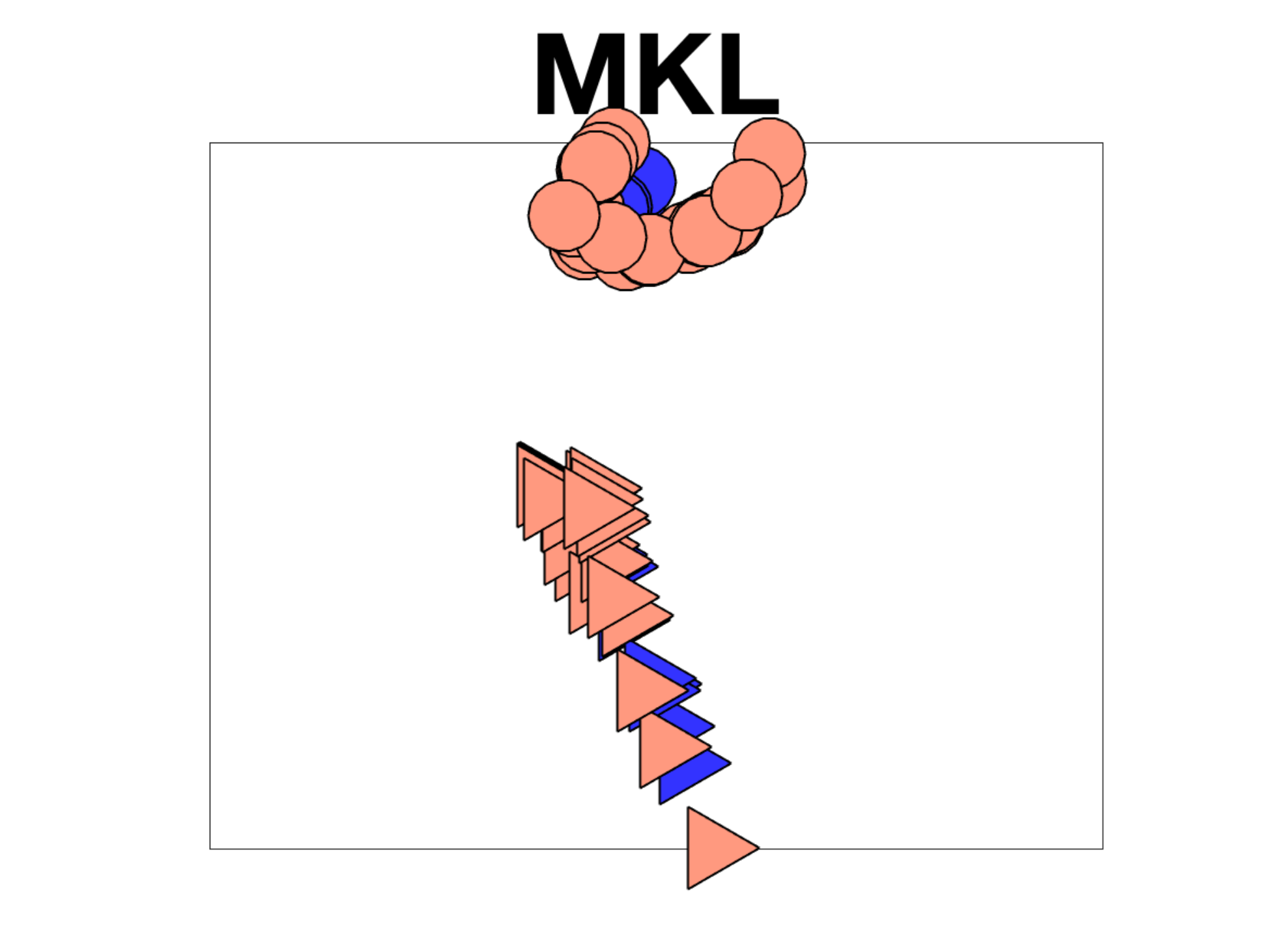}
\includegraphics[width=0.23\linewidth]{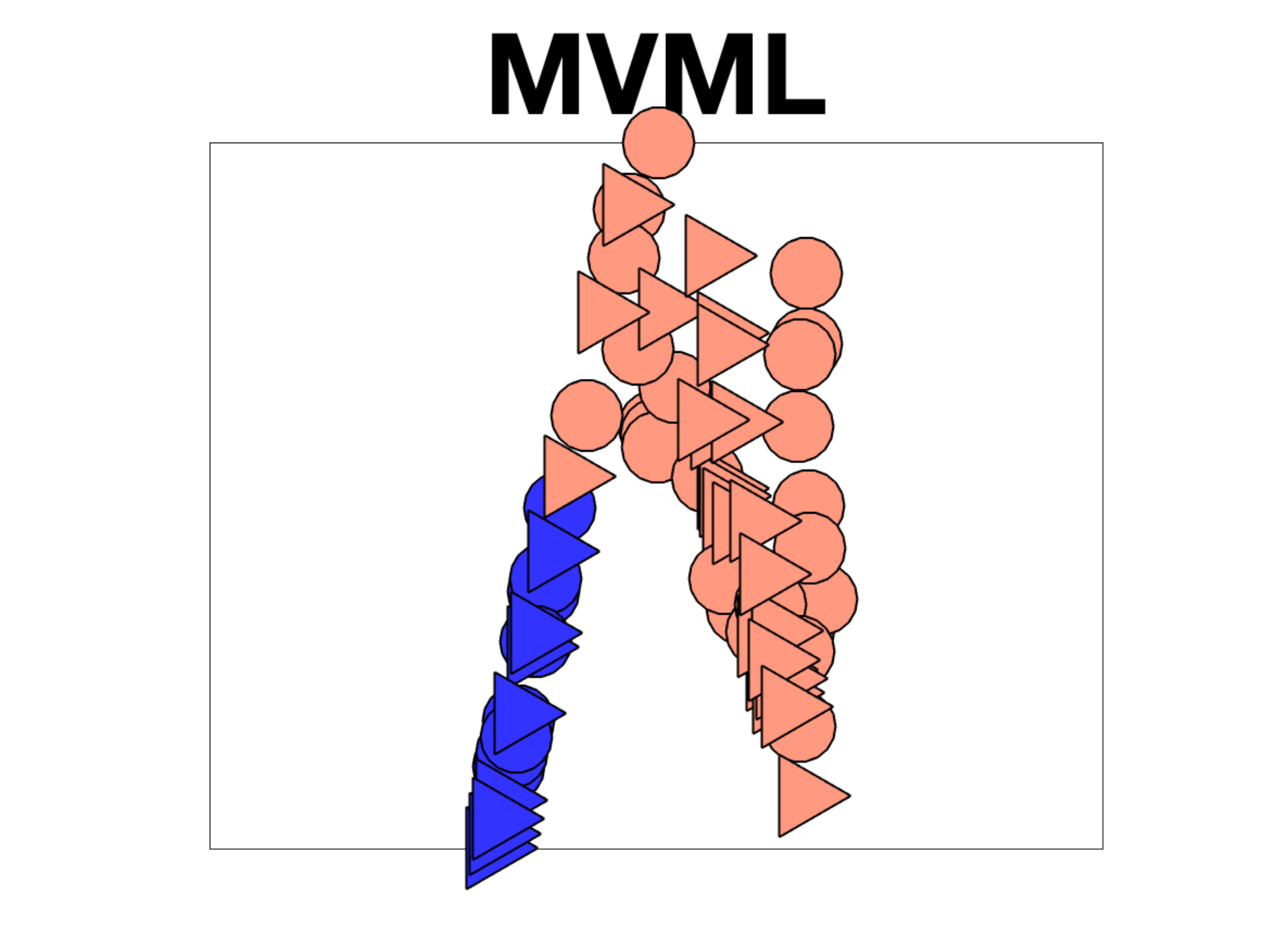}
\includegraphics[width=0.23\linewidth]{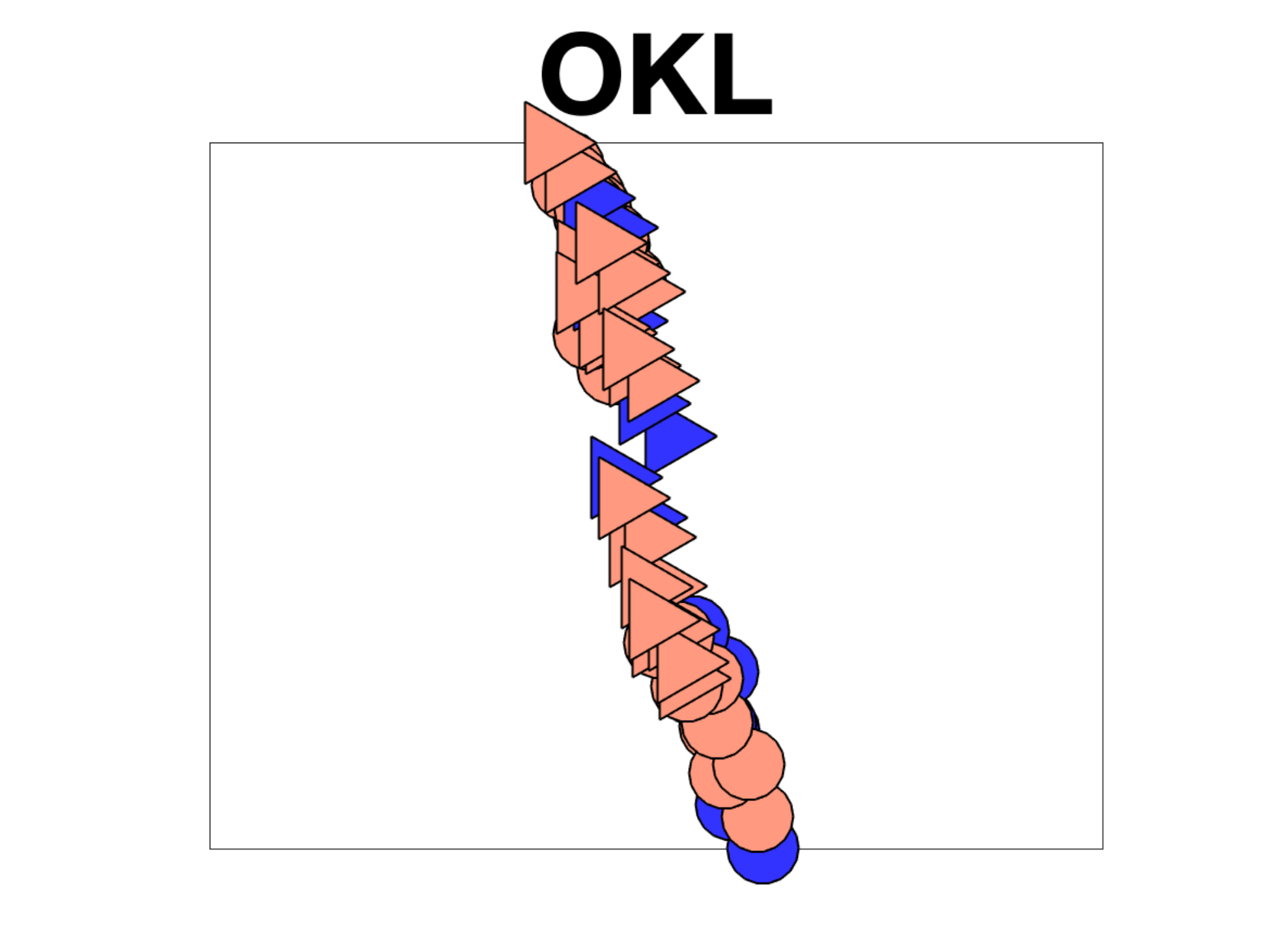}
\caption{Simple two-view dataset and its transformations - left: original data where one of the views is completely generated from the other by a linear transformation (a shear mapping followed by a rotation), left middle: MKL transformation, right middle: MVML transformation and right: OKL transformation. MVML shows a linear separation of classes (blue/pale red) of the views (circles/triangles), while MKL and OKL do not. }
\label{fig:toy}
\end{figure*}

\subsection{Illustration}

We illustrate with simple toy data the effects of learning both within- and between-view metrics. We compare our method, MVML, to MKL that considers only within-view dependencies, and to output kernel learning~(OKL) \cite{Ciliberto2015convex, dinuzzo2011learning} where separable kernels are learnt. 
We generated an extremely simple dataset of two classes and two views in $\mathbb{R}^2$, allowing for visualization and understanding of the way the methods perform classification with multi-view data.
The second view in the dataset is completely generated from the first, through a linear transformation~(a shear mapping followed by a rotation).
The generated data and transformation arising from applying the algorithms are shown in Figure~\ref{fig:toy}.
The space for transformed data is $\R^2$ since we used linear kernels for simplicity. Our MVML is the only method giving linear separation of the two classes. This means that it groups the data points into groups based on their class, not view, and thus is able 
to construct a good approximation of the initial data transformations by which we generated the second view.

\subsection{Rademacher complexity bound}
We now provide a generalization analysis of MVML algorithm using Rademacher complexities~\cite{bartlett2002rademacher}. The notion of Rademacher complexity has been generalizable to vector-valued hypothesis spaces~\cite{maurer2006rademacher, sindhwani2012scalable, sangnier2016joint}. 
Previous work has analyzed the case where the matrix-valued kernel is fixed prior to learning, while our analysis considers the kernel learning problem. It provides a Rademacher bound for our algorithm when both the vector-valued function $f$ and the metric between views $\mathbf{A}$ are learnt. 
We start by recalling that the feature map associated to the matrix-valued kernel $K$ is the mapping $\Gamma: \mathcal{X} \to \mathcal{L}(\mathcal{Y}, \mathcal{H})$, where $\mathcal{X}$ is the input space, $\mathcal{Y}=\mathbb{R}^v$, and $\mathcal{L}(\mathcal{Y}, \mathcal{H})$ is the set of bounded linear operators from $\mathcal{Y}$ to $\mathcal{H}$~(see, e.g., ~\cite{Micchelli2005onlearning,Carmeli2010vector} for more details). It is known that $K(x,z) = \Gamma(x)^*\Gamma(z)$. We denote by $\Gamma_\mathbf{A}$ the feature map associated to our multi-view kernel~(Equation~\ref{eq:mvka}).
The hypothesis class of MVML is 
\begin{equation*}
\mathcal{H}_\lambda = \{ x\mapsto f_{u,\mathbf{A}}(x) = \Gamma_\mathbf{A}(x)^* u : \mathbf{A}\in \Delta, \|u\|_{\mathcal{H}} \leq \beta\},
\end{equation*}
with $ \Delta = \{\mathbf{A}: \mathbf{A} \succ 0,\,\|\mathbf{A}\|_F \leq \alpha \}$ and $\beta$ is a regularization parameter. 
Let $\boldsymbol{\sigma}_1,\ldots,\boldsymbol{\sigma}_v$ be an iid family of vectors of independent Rademacher variables where $\boldsymbol{\sigma}_i\in \mathbb{R}^v, \; \forall\, i=1,\ldots,n$. The empirical Rademacher complexity of the vector-valued class $\mathcal{H}_\lambda$ is the function $\hat{\mathcal{R}}_n( \mathcal{H}_\lambda)$ defined as
\begin{equation*}
\hat{\mathcal{R}}_n( \mathcal{H}_\lambda) = \frac{1}{n} \E\left[\sup_{f\in\mathcal{H}} \sup_{\mathbf{A}\in \Delta}\sum_{i=1}^n \boldsymbol{\sigma}_i^\top f_{u,\mathbf{A}}(x_i)\right].
\end{equation*}
\begin{theo}\label{th:Rademacher}
The empirical Rademacher complexity of  $\mathcal{H}_\lambda$ can be upper bounded as follows:
\begin{equation*}
\hat{\mathcal{R}}_n( \mathcal{H}_\lambda) \leq \frac{\beta \sqrt{\alpha \|q\|_1}}{n},
\end{equation*}
where $q = \big(tr(\mathbf{K}_l^2)\big)_{l=1}^v$, and $\mathbf{K}_l$ is the Gram matrix computed from the training set $\{x_1,\ldots,x_n\}$ with the kernel $k_l$ defined on the view $l$. 
For kernels $k_l$ such that $tr(\mathbf{K}_l^2) \leq \tau n$, we have 
\begin{equation*}
\hat{\mathcal{R}}_n( \mathcal{H}_\lambda) \leq \beta\sqrt{\frac{ \alpha \tau v}{n}}.
\end{equation*}
\end{theo}
The proof for the theorem can be found in the supplementary material (Appendix A.2). Using well-known results~\cite[chapter 10]{mohri2012foundations}, this bound on Rademacher complexity can be used to obtain a generalization bound for our algorithm. It is worth mentioning that in our multi-view setting the matrix-valued kernel is computed from the product of the kernel matrices defined over the views. This is why our assumption is on the trace of the square of the kernel matrices $\mathbf{K}_l$. It is more restrictive than the usual one in the one-view setting~($tr(\mathbf{K}_l) \leq \tau n$), but is satisfied in some cases, like, for example, for diagonally dominant kernel matrices~\cite{scholkopf2002kernel}. It is interesting to investigate whether our Rademacher bound could be obtained under a much less restrictive assumption on the kernels over the views, and this will be investigated in future work.

\subsection{Block-sparsity and efficient implemen-\\tation via block-Nyström approximation}

In this section we consider variant of our formulation (\ref{eq:mvmlA}) which allows block-sparse solutions for the metric matrix~$\mathbf{A}$, and further show how to reduce the complexity of the required computations for our algorithms.

\paragraph{Block-sparsity}

We formulate a second optimization problem to study the effect of sparsity over $\mathbf{A}$. Instead of having for example $l_1$-norm regularizer over the whole matrix, we consider sparsity on a group level so that whole blocks corresponding to pairs of views are put to zero. 
Intuitively, the block-sparse result will give insight as to which views are interesting and worth taking into account in learning. For example, by tuning the parameter controlling sparsity level one could derive, in some sense, an order of importance to the views and their combinations. 
The convex optimization problem is as follows
\begin{align}
\label{eq:mvmlB}
 \min_{\mathbf{A},\mathbf{c}} \;\; &\| \mathbf{y}- (\mathbf{w}^T \otimes \mathbf{I}_n)\mathbf{H}\mathbf{AHc}\|^2 + \lambda\left\langle \mathbf{H}\mathbf{AHc},\mathbf{c} \right\rangle \notag \\
& + \eta \sum_{\gamma\in \mathcal{G}} \|\mathbf{A}_{\gamma}\|_F, 
\end{align} 
where we have a $l_1/l_2$-regularizer over set of groups $\mathcal{G}$ we consider for sparsity. In our multi-view setting these groups correspond to combinations of views; e.g. with three views the matrix $\mathbf{A}$ would consist of six groups: 

\hspace*{1cm}\begin{tabular}{ c c c c c c c }


\begin{tikzpicture}
\draw[step=0.2cm,black,thin] (0,0) grid (0.6,0.6);
\filldraw[fill=black!40!white, draw=black] (0,0.4) rectangle (0.2,0.6);
\end{tikzpicture}

&

\begin{tikzpicture}
\draw[step=0.2cm,black,thin] (0,0) grid (0.6,0.6);
\filldraw[fill=black!40!white, draw=black] (0.2,0.2) rectangle (0.4,0.4);
\end{tikzpicture}

&

\begin{tikzpicture}
\draw[step=0.2cm,black,thin] (0,0) grid (0.6,0.6);
\filldraw[fill=black!40!white, draw=black] (0.4,0.0) rectangle (0.6,0.2);
\end{tikzpicture}
 
&

\begin{tikzpicture}
\draw[step=0.2cm,black,thin] (0,0) grid (0.6,0.6);
\filldraw[fill=black!40!white, draw=black] (0,0.2) rectangle (0.2,0.4);
\filldraw[fill=black!40!white, draw=black] (0.2,0.4) rectangle (0.4,0.6);
\end{tikzpicture}

&

\begin{tikzpicture}
\draw[step=0.2cm,black,thin] (0,0) grid (0.6,0.6);
\filldraw[fill=black!40!white, draw=black] (0.0,0.0) rectangle (0.2,0.2);
\filldraw[fill=black!40!white, draw=black] (0.4,0.4) rectangle (0.6,0.6);
\end{tikzpicture}

&

\begin{tikzpicture}
\draw[step=0.2cm,black,thin] (0,0) grid (0.6,0.6);
\filldraw[fill=black!40!white, draw=black] (0.2,0.0) rectangle (0.4,0.2);
\filldraw[fill=black!40!white, draw=black] (0.4,0.2) rectangle (0.6,0.4);
\end{tikzpicture}

\end{tabular}

We note that when we speak of combinations of views we include both blocks of the matrix that this combination corresponds to. Using this group regularization, in essence, allows us to have view-sparsity in our multi-view kernel matrix.

To solve this optimization problem we introduce the same mapping as before, and obtain the same solution for $\mathbf{g}$ and $\mathbf{w}$.
However~(\ref{eq:mvmlB})  does not have an obvious closed-form solution for $\mathbf{A}$ so it is solved with proximal gradient method, the update rule being 
\begin{align}\label{eq:A_iter}
&[\mathbf{A}^{k+1}]_\gamma = \\
&\left( 1-\frac{\eta}{\|[\mathbf{A}^k -\mu^k \nabla h(\mathbf{A}^k)]_\gamma\|_F} \right)_+ [\mathbf{A}^k -\mu^k \nabla h(\mathbf{A}^k)]_\gamma, \notag
\end{align}
where $\mu^k$ is the step size, 
$h(\mathbf{A}^k) = \lambda\left\langle \mathbf{g},(\mathbf{A}^k)^\dagger \mathbf{g} \right\rangle$, and $\nabla h(\mathbf{A}^k) = -\lambda (\mathbf{A}^k)^{-1}\mathbf{g}\mathbf{g}^T(\mathbf{A}^k)^{-1}$. 

We note that even if we begin iteration with positive definite (pd) matrix the next iterate is not guaranteed to be always pd, and this is the reason for omitting the positivity constraint in the formulation of sparse problem~(Equation \ref{eq:mvmlB}). 
Nevertheless all block-diagonal results are pd, and so are other results if certain conditions hold. In experiments we have observed that the solution is positive semidefinite. The full derivation of the proximal algorithm and notes about positiveness of $\mathbf{A}$ are in supplementary material~(Appendix A.1).

\paragraph{Nyström approximation}

As a way to reduce the complexity of the required computations we propose using Nyström approximation on each one-view kernel matrix. 
In Nyström approximation method~\cite{williams2001using}, a~(scalar-valued) kernel matrix $\mathbf{M}$ is divided in four blocks, \[\mathbf{M} = \begin{bmatrix}
\mathbf{M}_{11} & \mathbf{M}_{12} \\
\mathbf{M}_{21} & \mathbf{M}_{22}
\end{bmatrix},\] 
and is approximated by $\mathbf{M} \approx \mathbf{Q}\mathbf{W}^\dagger \mathbf{Q}^T$, where  $\mathbf{Q}= \begin{bmatrix}
\mathbf{M}_{11} & \mathbf{M}_{12}
\end{bmatrix}^T$ and $\mathbf{W} = \mathbf{M}_{11}$. 
Denote $p$ as the number of rows of $\mathbf{M}$ chosen to build $\mathbf{W}$. This scheme gives a low-rank approximation of $\mathbf{M}$ by sampling $p$ examples, and only the last block, $\mathbf{M}_{22}$, will be approximated.

We could approximate the block kernel matrix $\mathbf{K}$ directly by applying the Nyström approximation, but this would have the effect of removing the block structure in the kernel matrix and consequently the useful multi-view information might be lost. 
Instead, we proceed in a way that is consistent with the multi-view problem and approximate each kernel matrix defined over one view as $\mathbf{K}_l \approx \mathbf{Q}_l\mathbf{W}_l^\dagger \mathbf{Q}_l^T = \mathbf{Q}_l (\mathbf{W}_l^\dagger)^{1/2}(\mathbf{W}_l^\dagger)^{1/2} \mathbf{Q}_l^T = \mathbf{U}_l\mathbf{U}_l^T , \forall\, l =1,\ldots,v$.  
The goodness of approximation is based on the $p$ chosen. Before performing the approximation a random ordering of the samples is calculated. 
We note that in our multi-view setting we have to impose the same ordering over all the views. 
We introduce the Nyström approximation to all our single-view kernels and define $\mathbf{U} = \text{blockdiag}{(\mathbf{U}_1,\cdots,\mathbf{U}_v)}$. 
We can now approximate our multi-view kernel (\ref{eq:mvkm}) 
as \[\mathbf{K} = \mathbf{HAH} \approx \mathbf{UU}^T\mathbf{AUU}^T = \mathbf{U}\tilde{\mathbf{A}}\mathbf{U}^T,\] 
where we have written $\tilde{\mathbf{A}} = \mathbf{U}^T\mathbf{AU}$.  Using this scheme, we obtain a block-wise Nyström approximation of $\mathbf{K}$ that preserves the multi-view structure of the kernel matrix while allowing substantial computational gains.

We introduce Nyström approximation into (\ref{eq:mvmlA}) and (\ref{eq:mvmlB}) and write $\tilde{\mathbf{g}} = \tilde{\mathbf{A}}\mathbf{U}^T\mathbf{c},$ resulting in  
\begin{align}
\label{eq:mvmlter}
\min_{\tilde{\mathbf{A}},\tilde{\mathbf{g}},\mathbf{w}} \;\; &\| \mathbf{y}- (\mathbf{w}^T \otimes \mathbf{I}_n)\mathbf{U}\tilde{\mathbf{g}}\|^2 + \lambda \langle \tilde{\mathbf{g}},\tilde{\mathbf{A}}^\dagger \tilde{\mathbf{g}} \rangle \\
&+ \eta\|\tilde{\mathbf{A}}\|_F^2, \;\;\;\; s.t. \;\; \tilde{\mathbf{A}} \succ 0 \notag
\end{align} 
and 
\begin{align} \label{eq:mvml_sparse_approx}
\min_{\tilde{\mathbf{A}},\tilde{\mathbf{g}},\mathbf{w}} \;\; &\| \mathbf{y}- (\mathbf{w}^T \otimes \mathbf{I}_n)\mathbf{U}\tilde{\mathbf{g}}\|^2 + \lambda \langle \tilde{\mathbf{g}},\tilde{\mathbf{A}}^\dagger \tilde{\mathbf{g}} \rangle \\
& + \eta \sum_{\gamma\in \mathcal{G}} \|\tilde{\mathbf{A}}_\gamma\|_F. \notag
\end{align}
We note that the optimization problems are not strictly equivalent to the ones before; namely we impose the Frobenius norm regularization over $\tilde{\mathbf{A}}$ rather than over $\mathbf{A}$. 
The obtained solution for~(\ref{eq:mvmlter}) will again satisfy the positivity condition when $\mu\eta<\tfrac{1}{2}$. For the sparse solution the positivity is unfortunately not always guaranteed, but is achieved if certain conditions hold. 

We solve the problems as before, and obtain: 
\begin{equation}\label{eq:D_approx}\tilde{\mathbf{g}} = (\mathbf{U}^T(\mathbf{w}^T \otimes \mathbf{I}_n)^T (\mathbf{w}^T \otimes \mathbf{I}_n)\mathbf{U}+\lambda \tilde{\mathbf{A}}^\dagger)^{-1}\mathbf{U}^T(\mathbf{w}^T \otimes \mathbf{I}_n)^T\mathbf{y}, \end{equation} 
\begin{equation}\label{eq:A_approx} 
\tilde{\mathbf{A}}^{k+1} = \left(1-2\mu\eta\right)\tilde{\mathbf{A}}^k + \mu\lambda(\tilde{\mathbf{A}}^k)^\dagger \tilde{\mathbf{g}}\tilde{\mathbf{g}}^T(\tilde{\mathbf{A}}^k)^\dagger,
\end{equation}
and
\begin{equation}\label{eq:w_approx}\mathbf{w} = (\tilde{\mathbf{Z}}^T\tilde{\mathbf{Z}})^{-1}\tilde{\mathbf{Z}}^T\mathbf{y}.
 \end{equation}
Here $\tilde{\mathbf{Z}}\in \R^{n\times v}$ is filled columnwise from $\mathbf{U}\tilde{\mathbf{g}}$.
For our block-sparse method with we get update rule 
\begin{align}\label{eq:A_approx_iter}
&[\tilde{\mathbf{A}}^{k+1}]_\gamma = \\
&\left( 1-\frac{\eta}{\left\|[\tilde{\mathbf{A}}^k -\mu^k \nabla f(\tilde{\mathbf{A}}^k)]_\gamma\right\|_F} \right)_+ \left[\tilde{\mathbf{A}}^k -\mu^k \nabla f(\tilde{\mathbf{A}}^k)\right]_\gamma, \notag
\end{align}
where 
$\nabla f(\tilde{\mathbf{A}}^k) = -\lambda (\tilde{\mathbf{A}}^k)^{-1}\mathbf{g}\mathbf{g}^T(\tilde{\mathbf{A}}^k)^{-1}$  and $\mu^k$ is the step size.


\begin{figure*}[tb]
\centering
\includegraphics[width=0.96\textwidth]{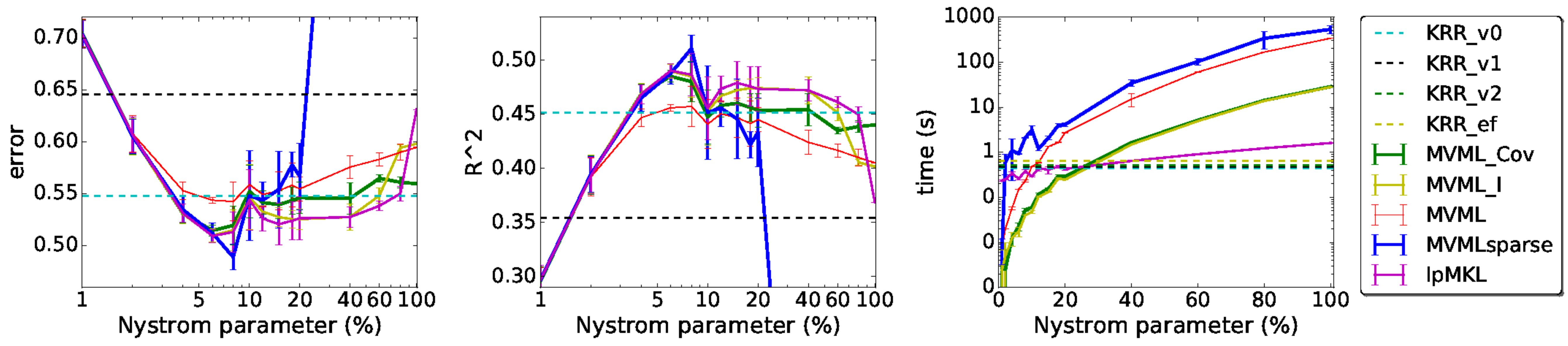}
\caption{Regression on Sarcos1-dataset. Left: normalized mean squared errors (the lower the better), middle: $R^2$-score (the higher the better), right: running times; as functions of Nyström approximation level (note the logartithmic scales). The results for KRR are calculated without approximation and are shown as horizontal dashed lines. Results for view 2 and early fusion are worse than others and outside of the scope of the two plots. } \label{fig:sarcos}
\end{figure*}

We follow the same algorithm than before for calculating the solution;  now over $\tilde{\mathbf{A}}$ and $\tilde{\mathbf{g}}$ 
(Algorithm~\ref{alg:mvml}, version b). 
The complexity is now of order $\mathcal{O}(v^3p^3)$ rather than $\mathcal{O}(v^3n^3)$, where $p\ll n$ is the number of samples chosen for the Nyström approximation in each block. 
From the obtained solution it is possible to calculate the original $\mathbf{g}$ and $\mathbf{A}$ if needed.

To also reduce the complexity of predicting with our multi-view kernel framework, our block-wise Nyström approximation is used again on the test kernel matrices $\mathbf{K}_s^{test}$ computed with the test examples. 
Let us recall that for each of our single-view kernels, we have an approximation $\mathbf{K} \approx \mathbf{U}_s\mathbf{U}_s^T = \mathbf{Q}_s\mathbf{W}_s^\dagger \mathbf{Q}_s^T$. We choose $\mathbf{Q}_s^{test}$ to be $p$ first columns of the matrix $\mathbf{K}_s^{test}$, and define the approximation for the test kernel to be \[ \mathbf{K}_s^{test} \approx \mathbf{Q}_s^{test}\mathbf{W}_s^\dagger \mathbf{Q}_s^T = \mathbf{Q}_s^{test} \left( \mathbf{W}_s^\dagger \right)^{1/2} \mathbf{U}_s^{T}. \]
In such an approximation, the error is in the last $n-p$ columns of $\mathbf{K}_s^{test}$. 
We gain in complexity, as if forced to use the test kernel as is, we would need to calculate $\mathbf{A}$ from $\tilde{\mathbf{A}}$ in $\mathcal{O}(vn^3)$ operations.

\section{Experiments}


Here we evaluate the proposed multi-view metric learning~(MVML) method on real-world datasets and compare it to relevant methods. 
The chosen datasets are "pure" multi-view datasets, that is to say, the view division arises naturally from the data. 

We perform two sets of experiments with two goals. First, we evaluate our method in regression setting with a large range of Nyström approximation levels in order to understand the effect it has on our algorithm. Secondly, we compare MVML to relevant state-of-the-art methods in classification. In both cases, we use non multi-view methods to justify the multi-view approach.
The methods we use in addition to our own are:
\vspace*{-0.45cm}\begin{itemize} \setlength\itemsep{-0.5em}

\item \textbf{MVML\_Cov and MVML\_I}: we use pre-set kernels in our framework: MVML\_Cov uses the kernel from~\cite{Kadri2013multi} and MVML\_I refers to the case when have only one-view kernel matrices in the diagonal of the multi-view kernel.\footnote{Code for MVML is available at https://lives.lif.univ-mrs.fr/?page\_id=12}
\item \textbf{lpMKL} is an algorithm for learning weights for MKL kernel~\cite{kloft2011lp}. We apply it to kernel regression.
\item \textbf{OKL}~\cite{Ciliberto2015convex, dinuzzo2011learning} is a kernel learning method for separable kernels.
\item \textbf{MLKR}~\cite{Weinberger2007metric} is an algorithm for metric learnig in kernel setting.
\item \textbf{KRR and SVM}: We use kernel ridge regression and support vector machines with one-view as well as in early fusion~(ef) and late fusion~(lf) in order to validate the benefits of using multi-view~methods.
\end{itemize}
\vspace*{-0.45cm}We perform our experiments with Python, but for OKL and MLKR we use the MATLAB codes provided by authors\footnote{https://www.cs.cornell.edu/{\mytilde}kilian/code/code.html}\myfootnote{and https://github.com/cciliber/matMTL.}\hspace{-0.65cm}.
In MVML we 
set weights uniformly to $\tfrac{1}{v}$. For all the datasets we use Gaussian kernels, $k(\mathbf{x}, \mathbf{z}) = exp(-\tfrac{1}{2\sigma^2}\|\mathbf{x}-\mathbf{z}\|^2)$.

\begin{table*}[ht]
\caption{Classification accuracies $\!\pm\!$ standard deviation.~The number after the dataset indicates the level of approximation of the kernel.~The results for efSVM classification for Flower17-dataset are missing as only~similarity matrices for each view were provided. Last column reports the best result obtained when using only one view.} \label{tbl:classification_results}
\begin{center}
\resizebox{\textwidth}{!}{%
\begin{tabular}{llllllll}
\\
{\bf METHOD}  &{\bf MVML}  &{\bf MVMLsp.}  &{\bf MVML\_Cov}  &{\bf MVML\_I}  &{\bf lpMKL}  &{\bf OKL} &{\bf MLKR}  \\
\hline \\
Flower17 (6\%) & 75.98 $\pm$ 2.62 & 75.71 $\pm$ 2.48 & 75.71 $\pm$ 2.19 & \textbf{76.03 $\pm$ 2.36} & 75.54 $\pm$ 2.61 & 68.73 $\pm$ 1.95 & 63.82 $\pm$ 2.51  \\  %
Flower17 (12\%) & 77.89 $\pm$ 2.41 & 77.43 $\pm$ 2.44 & 77.30 $\pm$ 2.36 & \textbf{78.36 $\pm$ 2.52} & 77.87 $\pm$ 2.52 & 75.19 $\pm$ 1.97 & 64.41 $\pm$ 2.41 \\  %
Flower17 (24\%) & 78.60 $\pm$ 1.41 & 78.60 $\pm$ 1.36 & 79.00 $\pm$ 1.75 & \textbf{79.19 $\pm$ 1.51} & 78.75 $\pm$ 1.58 & 76.76 $\pm$ 1.62 & 65.44 $\pm$ 1.36 \\  %
uWaveG. (6\%) & 92.67 $\pm$ 0.21 & \textbf{92.68 $\pm$ 0.17} & 92.34 $\pm$ 0.20 & 92.34 $\pm$ 0.19 & 92.34 $\pm$ 0.18 & 70.09  $\pm$ 1.07 & 71.09 $\pm$ 0.94  \\  %
uWaveG. (12\%) & \textbf{93.03 $\pm$ 0.11} & 92.86 $\pm$ 0.26 & 92.53 $\pm$ 0.18 & 92.59 $\pm$ 0.13 & 92.48 $\pm$ 0.21 & 74.07 $\pm$ 0.26 & 80.22 $\pm$ 0.38  \\  %
uWaveG. (24\%) & 92.59 $\pm$ 0.99 & \textbf{93.26 $\pm$ 0.15} & 92.66 $\pm$ 0.05 & 93.10 $\pm$ 0.11 & 92.85 $\pm$ 0.13 & 76.65 $\pm$ 0.33 & 86.38  $\pm$ 0.31  \\  %
\\
& & {\bf METHOD}  &{\bf efSVM} &{\bf lfSVM} &{\bf 1 view SVM} &  &  \\
\cline{3-6} \\
& & Flower17 (6\%) & \multicolumn{1}{c}{-} & 15.32 $\pm$ 1.94 & 11.59 $\pm$ 1.54  &  &   \\ 
& & Flower17 (12\%) & \multicolumn{1}{c}{-} & 23.82 $\pm$ 2.38 & 15.74 $\pm$ 1.54  &  &   \\ 
& & Flower17 (24\%) & \multicolumn{1}{c}{-} & 38.24 $\pm$ 2.31 & 22.79 $\pm$ 0.79   &   & \\
& &  uWaveG. (6\%) & 80.00 $\pm$ 0.74 & 71.24 $\pm$ 0.41 & 56.54 $\pm$ 0.38 &  &  \\ 
& & uWaveG. (12\%) & 82.29 $\pm$ 0.63 & 72.53 $\pm$ 0.16 & 57.50 $\pm$ 0.17  &  & \\  
& & uWaveG. (24\%) & 84.07 $\pm$ 0.23 & 72.99 $\pm$ 0.06 & 58.01 $\pm$ 0.05  &  &   

\end{tabular}}
\end{center}

\end{table*}

\subsection{Effect of Nyström approximation}

For our first experiment we consider SARCOS-dataset\footnote{http://www.gaussianprocess.org/gpml/data.}, where the task is to map a 21-dimensional input space~(7 joint positions, 7 joint velocities, 7 joint accelerations) to the corresponding 7 joint torques. Here we present results to the first task. 

The results with various levels of Nyström approximation - averaged over four approximations - from 1\% to 100\% of data are shown in Figure \ref{fig:sarcos}. Regularization parameters were cross-validated over values $\lambda\in[1$e-$08,10]$ and $\eta\in[1$e-$04,100]$. Kernel parameter $\gamma = 1/2\sigma^2$ was fixed to be 1/number of features as a trade-off between overfitting and underfitting. We used only 1000 data samples of the available 44484 in training (all 4449 in testing) to be feasibly able to show the effect of approximating the matrices on all levels, and wish to note that using more data samples with moderate approximation level we can yield a lower error than presented here: for example with 2000 training samples and Nyström approximation level of 8\% we obtain error of 0.3915. 
However the main goal of our experiment was to see how our algorithm behaves with various Nyström approximation levels and because of the high complexity of our algorithm trained on the full dataset without approximation we performed this experiment with low amount of samples.

The lowest error was obtained with our MVMLsparse algorithm at 8\% Nyström approximation level. 
All the multi-view results seem to benefit from using the approximation. Indeed, approximating the kernel matrices can be seen as a form of regularization and our results reflect on that~\cite{rudi2015less}. 
Overall our MVML learning methods have much higher computational cost with large Nyström parameters, as can be seen from Figure~\ref{fig:sarcos}, rightmost plot. However with smaller approximation levels with which the methods are intended to be used, the computing time is competitive.

\subsection{Classification results}

In our classification experiments we use 
two real-world multi-view datasets: Flower17\footnote{http://www.robots.ox.ac.uk/{\mytilde}vgg/data/flowers/17.}~(7 views, 17 classes, 80 samples per class) and uWaveGesture\footnote{http://www.cs.ucr.edu/{\mytilde}eamonn/time\_series\_data.}~(3 views, 8 classes, 896 data samples for training and 3582 samples for testing). 
We set the kernel parameter to be mean of distances, $\sigma = \tfrac{1}{n^2}\sum_{i,j=1}^n \|\mathbf{x}_i-\mathbf{x}_j\| $.
The regularization parameters were obtained by cross-validation over values $\lambda\!\in\!\!\,[1$e-$08,10]$ and $\eta\in[1$e-$03,100]$. The results are averaged over four approximations. 

We adopted one-vs-all classification approach for multiclass classification.
The results are displayed in Table~\ref{tbl:classification_results}. 
The MVML results are always notably better than the SVM results, or the results obtained with OKL or MLKR. Compared to MVML, OKL and MLKR accuracies decrease more with low approximation levels. 
We can see that all MVML methods perform very similarly, sometimes the best result is obtained with fixed multi-view kernel, sometimes when $\mathbf{A}$ is learned.

\begin{figure}[h]
\centering
\includegraphics[width=0.47\linewidth]{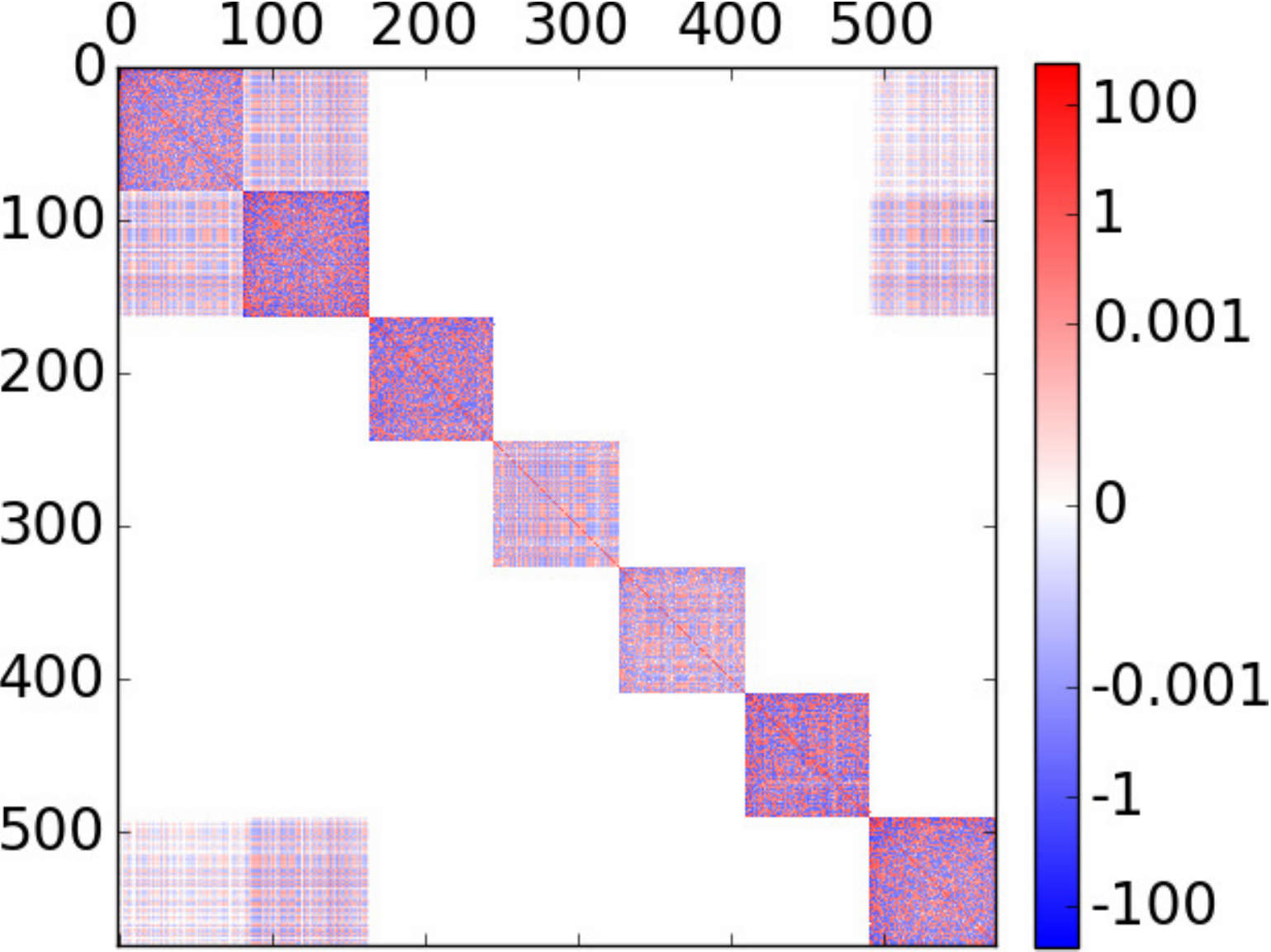}
\caption{An example of learned $\tilde{\mathbf{A}}$ with MVMLsparse from Flower17~(12\%) experiments.} \label{fig:flower_sparse_A}
\end{figure}

As an example of our sparse output with MVML we note that running the algorithm with Flower17 dataset with 12\% approximation often resulted in a spd matrix as in Figure \ref{fig:flower_sparse_A}. Indeed the resulting sparsity is very interesting and tells us about importance of the views and their interactions.

\section{Conclusion}

We have introduced a general class of matrix-valued multi-view kernels for which we have presented two methods for simultaneously learning a multi-view function and a metric in vector-valued kernel spaces.
We provided iterative algorithms for the two resulting optimization problems, and have been able to significantly lower the high computational cost associated with kernel methods by introducing block-wise Nyström approximation.
We have explained the feasibility of our approach onto a trivial dataset which reflects the objective of learning the within-view and between-view correlation metrics.
The performance of our approach was illustrated with experiments with real multi-view datasets by comparing our method to standard multi-view approaches, as well as methods for metric learning and kernel learning. Our sparse method is especially promising in the sense that it could give us information about importance of the views.
It would be interesting to investigate the applicability of our framework in problems involving missing data in views, as well as the generalization properties with the Nyström approximation. We would also like to continue investigating the theoretical properties of our sparse algorithm in order to prove the positiveness of the learned metric matrix that we observed experimentally.

\section*{Acknowledgements}

We thank the anonymous reviewers for their relevant and helpful comments. 
This work is granted by Lives Project (ANR-15-CE23-0026).

\bibliographystyle{plain}
\bibliography{mvml}

\clearpage
\appendix
\thispagestyle{empty}
\twocolumn[

\aistatstitle{Multi-view Metric Learning in Vector-valued Kernel Spaces  {\Large Supplementary Material}}

\aistatsauthor{ Riikka Huusari \And Hachem Kadri \And  Cécile Capponi }
\vspace*{0.1cm}
\aistatsaddress{ Aix Marseille Univ, Université de Toulon, CNRS, LIS, Marseille, France  } ]

\section{Appendix}

\subsection{MVML optimization}

Here we go through the derivations of the solutions $\textbf{A}$, $\textbf{D}$ and $\textbf{w}$ for our optimization problem. The presented derivations are for the case without Nyström approximation; however the derivations with Nyström approximation are done exactly the same way.

\subsubsection*{Solving for $\mathbf{g}$ and $\mathbf{w}$}

Let us first focus on the case where $\textbf{A}$ and $\textbf{w}$ are fixed and we solve for $\textbf{g}$. We calculate the derivative of the expression in Equation~(\ref{eq:mvmlbis}): \begin{align*}
&\frac{d}{d\textbf{g}}\; \| \mathbf{y}- (\mathbf{w}^T \otimes \mathbf{I}_n)\mathbf{Hg}\|^2 + \lambda\left\langle \mathbf{g},\mathbf{A}^\dagger \mathbf{g} \right\rangle \\
&=\frac{d}{d\textbf{g}}\; \langle \mathbf{y},\mathbf{y} \rangle -2 \langle \mathbf{y}, (\mathbf{w}^T \otimes \mathbf{I}_n)\mathbf{Hg} \rangle  \\
&\;\;\;\;+ \langle (\mathbf{w}^T \otimes \mathbf{I}_n)\mathbf{Hg},(\mathbf{w}^T \otimes \mathbf{I}_n)\mathbf{HD} \rangle + \lambda \langle \mathbf{g},\mathbf{A}^\dagger \mathbf{g}  \rangle \\
&= -2\mathbf{H}(\mathbf{w}^T \otimes \mathbf{I}_n)^T\mathbf{y} \\
&\;\;\;\;+ 2\mathbf{H}(\mathbf{w}^T \otimes \mathbf{I}_n)^T(\mathbf{w}^T \otimes \mathbf{I}_n)\mathbf{Hg} +2\lambda \mathbf{A}^\dagger \mathbf{g}
\end{align*}
By setting this to zero we obtain the solution \[ \mathbf{g} = (\mathbf{H}(\mathbf{w}^T \otimes \mathbf{I}_n)^T(\mathbf{w}^T \otimes \mathbf{I}_n)\mathbf{H}+\lambda \mathbf{A}^\dagger)^{-1}\mathbf{H}(\mathbf{w}^T \otimes \mathbf{I}_n)^T\mathbf{y}. \]

As for $\mathbf{w}$ when $\textbf{A}$ and $\textbf{g}$ are fixed, we need only to consider optimizing \begin{equation}
\min_{\mathbf{w}} \;\; \| \mathbf{y}- (\mathbf{w}^T \otimes \mathbf{I}_n)\mathbf{Hg}\|^2.
\end{equation}
If we denote that $\mathbf{Z}\in\R^{n\times v}$ is equal to reshaping $\mathbf{Hg}$ by taking the elements of the vector and arranging them onto the columns of $\mathbf{Z}$, we obtain a following form: 
\begin{equation}
\min_{\mathbf{w}} \;\; \| \mathbf{y}- \mathbf{Z}\mathbf{w}\|^2.
\end{equation}
One can easily see by taking the derivative and setting it to zero that the solution for this is 
\begin{equation}
\mathbf{w} = \left( \mathbf{Z}^T \mathbf{Z}\right)^{-1} \mathbf{Z}^T \mathbf{y}.
\end{equation}

\subsection*{Solving for $\mathbf{A}$ in (\ref{eq:mvmlA})}

When we consider $\mathbf{g}$ (and $\mathbf{w}$) to be fixed in the MVML framwork~(\ref{eq:mvmlA}), for $\mathbf{A}$ we have the following minimization problem: \begin{equation*}
\min_{\mathbf{A}} \;\; \lambda\left\langle \mathbf{g},\mathbf{A}^\dagger \mathbf{g} \right\rangle + \eta\|\mathbf{A}\|_F^2 
\end{equation*} 
Derivating this with respect to $\mathbf{A}$ gives us 
\begin{align*}
&\frac{d}{d\textbf{A}}\; \lambda\left\langle \mathbf{g},\mathbf{A}^\dagger \mathbf{g} \right\rangle + \eta\|\mathbf{A}\|_F^2 \\
&=\frac{d}{d\textbf{A}}\; \lambda\left\langle \mathbf{g},\mathbf{A}^\dagger \mathbf{g} \right\rangle + \eta\; tr(\mathbf{A}\mathbf{A}) \\
&= -\lambda \mathbf{A}^\dagger \mathbf{gg}^T \mathbf{A}^\dagger\text{ \footnotemark} + 2\eta\mathbf{A} 
\end{align*} \footnotetext{\label{note_mtrxcb}Matrix cookbook (Equation 61): \url{https://www.math.uwaterloo.ca/~hwolkowi/matrixcookbook.pdf}.}
Thus the gradient descent step will be 
\[\mathbf{A}^{k+1} = \left(1-2\mu\eta\right)\mathbf{A}^k + \mu\lambda\left(\mathbf{A}^k\right)^\dagger \mathbf{gg}^T\left(\mathbf{A}^k\right)^\dagger\]
when moving to the direction of negative gradient with step size $\mu$.


\subsection*{Solving for $\mathbf{A}$ in (\ref{eq:mvmlB})}

To solve $\mathbf{A}$ from equation~(\ref{eq:mvmlB}) we use proximal minimization. Let us recall the optimization problem after the change of the variable: 
\begin{align*} \min_{\mathbf{A},\mathbf{g},\mathbf{w}} \;\; &\| \mathbf{y}- (\mathbf{w}^T \otimes \mathbf{I}_n)\mathbf{H}\mathbf{g}\|^2 + \lambda \langle \mathbf{g},\mathbf{A}^\dagger \mathbf{g} \rangle \\
& + \eta \sum_{\gamma\in \mathcal{G}} \|\mathbf{A}_\gamma\|_F, \end{align*}
and denote \[h(\mathbf{A}) = \lambda\left\langle \mathbf{g},\mathbf{A}^\dagger \mathbf{g} \right\rangle\] and \[\Omega(\mathbf{A}) = \eta \sum_{\gamma\in \mathcal{G}} \|\mathbf{A}_\gamma\|_F  \] for the two terms in our optimization problem that contain the matrix $\mathbf{A}$. 

Without going into detailed theory of proximal operators and proximal minimization, we remark that the proximal minimization algorithm update takes the form
\[ \mathbf{A}^{k+1} = \textbf{prox}_{\mu^k \Omega} (\mathbf{A}^k -\mu^k \nabla h(\mathbf{A}^k)). \]
It is well-known that in traditional group-lasso situation the proximal operator is
\[ [ \textbf{prox}_{\mu^k \Omega} (\mathbf{z}) ]_\gamma = \left( 1-\frac{\eta}{\|\mathbf{z}_\gamma\|_2} \right)_+ \mathbf{z}_\gamma, \] where $\mathbf{z}$ is a vector and $+$ denotes the maximum of zero and the value inside the brackets. In our case we are solving for a matrix, but due to the equivalence of Frobenious norm to vector 2-norm we can use this exact same operator. Thus we get as the proximal update 
\begin{align*}
&[\mathbf{A}^{k+1}]_\gamma = \\
&\left( 1-\frac{\eta}{\|[\mathbf{A}^k -\mu^k \nabla h(\mathbf{A}^k)]_\gamma\|_F} \right)_+ [\mathbf{A}^k -\mu^k \nabla h(\mathbf{A}^k)]_\gamma,
\end{align*}
where 
\[\nabla h(\mathbf{A}^k) = -\lambda (\mathbf{A}^k)^{-1}\mathbf{g}\mathbf{g}^T(\mathbf{A}^k)^{-1}.\] %

We can see from the update fromula and the derivative that if $\mathbf{A}^k$ is a positive matrix, the update without block-multiplication, $\mathbf{A}^k -\mu^k \nabla h(\mathbf{A}^k)$, will be positive, too. This is unfortunately not enough to guarantee the general positivity of $\mathbf{A}^{k+1}$. However we note that it is, indeed, positive if it is block-diagonal, and in general whenever a matrix of the multipliers $\alpha$ \[\alpha_{st} = \left( 1-\frac{\eta}{\|[\mathbf{A}^k -\mu^k \nabla h(\mathbf{A}^k)]_{st}\|_2} \right)_+\] is positive, then $\mathbf{A}^{k+1}$ is, too (see~\cite{gunther2012schur} for reference - this is a blockwise Hadamard product where the blocks commute).

\subsection{Proof of Theorem~\ref{th:Rademacher}}

\begin{customthm}{1}\label{th:Rademacher_again}
Let $\mathcal{H}$ be a vector-valued RKHS associated with the the multi-view kernel $K$ defined by Equation~\ref{eq:mvka}. Consider the hypothesis class $\mathcal{H}_\lambda = \{ x\mapsto f_{u,\mathbf{A}}(x) = \Gamma_\mathbf{A}(x)^* u : \mathbf{A}\in \Delta, \|u\|_{\mathcal{H}} \leq \beta\}$, with $ \Delta = \{\mathbf{A}: \mathbf{A} \succ 0,\,\|\mathbf{A}\|_F \leq \alpha \}$.  
The empirical Rademacher complexity of  $\mathcal{H}_\lambda$ can be upper bounded as follows:
\begin{equation*}
\hat{\mathcal{R}}_n( \mathcal{H}_\lambda) \leq \frac{\beta \sqrt{\alpha \|q\|_1}}{n},
\end{equation*}
where $q = \big(tr(\mathbf{K}_l^2)\big)_{l=1}^v$, and $\mathbf{K}_l$ is the Gram matrix computed from the training set $\{x_1,\ldots,x_n\}$ with the kernel $k_l$ defined on the view $l$. 
For kernels $k_l$ such that $tr(\mathbf{K}_l^2) \leq \tau n$, we have 
\begin{equation*}
\hat{\mathcal{R}}_n( \mathcal{H}_\lambda) \leq \beta\sqrt{\frac{ \alpha \tau v}{n}}.
\end{equation*}
\end{customthm}
%

\textit{Proof.}
We start by recalling that the feature map associated to the operator-valued kernel $K$ is the mapping $\Gamma: \mathcal{X} \to \mathcal{L}(\mathcal{Y}, \mathcal{H})$, where $\mathcal{X}$ is the input space, $\mathcal{Y}=\mathbb{R}^v$, and $\mathcal{L}(\mathcal{Y}, \mathcal{H})$ is the set of bounded linear operators from $\mathcal{Y}$ to $\mathcal{H}$ (see, e.g., ~\cite{Micchelli2005onlearning,Carmeli2010vector} for more details). It is known that $K(x,z) = \Gamma(x)^*\Gamma(z)$. We denote by $\Gamma_\mathbf{A}$ the feature map associated to our multi-view kernel~(Equation~\ref{eq:mvka}). We also define the matrix $\boldsymbol{\Sigma} = (\boldsymbol{\sigma})_{i=1}^n \in \mathbb{R}^{nv}$

\begin{align*}
\hat{\mathcal{R}}_n( \mathcal{H}_\lambda) &= \frac{1}{n} \E\left[\sup_{f\in\mathcal{H}} \sup_{\mathbf{A}\in \Delta}\sum_{i=1}^n \boldsymbol{\sigma}_i^\top f_{u,\mathbf{A}}(x_i)\right] \\
&= \frac{1}{n} \E\left[\sup_{u} \sup_{\mathbf{A}}\sum_{i=1}^n \langle\boldsymbol{\sigma}_i , \Gamma_\mathbf{A}(x_i)^* u\rangle_{\mathbb{R}^v}\right]\\
& =  \frac{1}{n} \E\left[\sup_{u} \sup_{\mathbf{A}}\sum_{i=1}^n \langle \Gamma_\mathbf{A}(x_i) \boldsymbol{\sigma}_i ,  u\rangle_{\mathcal{H}}\right] \text{\quad (1)}\\
& \leq \frac{\beta}{n}  \E\left[ \sup_{\mathbf{A}}\|\sum_{i=1}^n  \Gamma_\mathbf{A}(x_i) \boldsymbol{\sigma}_i\|_{\mathcal{H}}\right] \text{\quad (2)}\\
& = \frac{\beta}{n}  \E\left[ \sup_{\mathbf{A}} \left(\sum_{i,j=1}^n \langle \boldsymbol{\sigma}_i,  K_{\mathbf{A}}(x_i,x_j) \boldsymbol{\sigma}_j\rangle_{\mathbb{R}^v}\right)^{\tfrac{1}{2}}\right] \text{\quad (3)}\\
& =  \frac{\beta}{n}  \E\left[ \sup_{\mathbf{A}} \left( \langle \boldsymbol{\Sigma},  \mathbf{K}_{\mathbf{A}}\boldsymbol{\Sigma}\rangle_{\mathbb{R}^{nv}}\right)^{1/2}\right] \\
& = \frac{\beta}{n}  \E\left[ \sup_{\mathbf{A}}  \langle \boldsymbol{\Sigma},  \mathbf{HAH}\boldsymbol{\Sigma}\rangle^{1/2}\right] \\
& = \frac{\beta}{n}  \E\left[ \sup_{\mathbf{A}} tr(\mathbf{H} \boldsymbol{\Sigma}\boldsymbol{\Sigma}^\top\mathbf{H}\mathbf{A})^{1/2}\right] \\
& \leq \frac{\beta}{n}  \E\left[ \sup_{\mathbf{A}} tr([\mathbf{H} \boldsymbol{\Sigma}\boldsymbol{\Sigma}^\top\mathbf{H}]^2)^{1/4} tr(\mathbf{A}^2)^{1/4}\right] \text{\quad (4)} \\
& \leq \frac{\beta}{n}  \E\left[ \sup_{\mathbf{A}} tr(\mathbf{H}^2 \boldsymbol{\Sigma}\boldsymbol{\Sigma}^\top)^{1/2} tr(\mathbf{A}^2)^{1/4}\right]\\
& \leq \frac{\beta\sqrt{\alpha}}{n}  \E\left[ \sup_{\mathbf{A}} tr(\mathbf{H}^2 \boldsymbol{\Sigma}\boldsymbol{\Sigma}^\top)^{1/2} \right]\\
& = \frac{\beta\sqrt{\alpha}}{n}  \E\left[tr(\mathbf{H}^2 \boldsymbol{\Sigma}\boldsymbol{\Sigma}^\top)^{1/2} \right] \\
& \leq \frac{\beta\sqrt{\alpha}}{n}  \left(\E\left[tr(\mathbf{H}^2 \boldsymbol{\Sigma}\boldsymbol{\Sigma}^\top)\right]\right)^{1/2}  \text{\quad (5)} \\
& = \frac{\beta\sqrt{\alpha}}{n}  \left(tr\left[\mathbf{H}^2\E(\boldsymbol{\Sigma}\boldsymbol{\Sigma}^\top)\right]\right)^{1/2}   \\
& =  \frac{\beta\sqrt{\alpha}}{n}  \sqrt{\|(tr(\mathbf{K_1}^2),\ldots, tr(\mathbf{K_v}^2))\|_1}.
\end{align*}
Here (1) and (3) are obtained with reproducing property, (2) and (4) with Cauchy-Schwarz inequality, and (5) with Jensen's inequality.
The last equality follows from the fact that $tr(\mathbf{H}^2) = \sum_{l=1}^v tr(\mathbf{K_l}^2)$.
For kernels $k_l$ that satisfy $tr(\mathbf{K}_l^2) \leq \tau n$, $l=1,\ldots, v$, we obtain that 
$$\hat{\mathcal{R}}_n( \mathcal{H}_\lambda) \leq \beta\sqrt{\frac{ \alpha \tau v}{n}}.  \;\;\; _\square $$
%

\end{document}